\documentclass[11pt,a4paper]{article}
\usepackage[hyperref]{acl2018}
\usepackage{times}
\usepackage{latexsym}

\usepackage{url}
\usepackage{helvet}
\usepackage{courier}
\usepackage{amsmath}
\usepackage{booktabs}
\usepackage{bm}
\usepackage{graphicx}
\usepackage{subfigure}
\usepackage{multirow}
\usepackage{array} 
\usepackage[normalem]{ulem}
\usepackage{bigdelim}

\usepackage{CJKutf8}          

\aclfinalcopy 
\def\aclpaperid{68} 

\title{Learning to Ask Questions in Open-domain Conversational Systems with Typed Decoders}

\author{Yansen Wang$^{1,}$\thanks{~~Authors contributed equally to this work.} , Chenyi Liu$^{1,*}$, Minlie Huang$^{1,}$\thanks{~~Corresponding author: Minlie Huang.} , Liqiang Nie$^2$\\
  $^1$Conversational AI group, AI Lab., Department of Computer Science, Tsinghua University\\
    $^1$Beijing National Research Center for Information Science and Technology, China\\
  $^2$Shandong University \\
  {\tt ys-wang15@mails.tsinghua.edu.cn;liucy15@mails.tsinghua.edu.cn;}\\
  {\tt aihuang@tsinghua.edu.cn;nieliqiang@gmail.com}}

\date{}

\begin{document}
\begin{CJK*}{UTF8}{gkai}

\maketitle
\begin{abstract}
Asking good questions in large-scale, open-domain conversational systems is quite significant yet rather untouched. 
This task, substantially different from traditional question generation, requires to question not only with various patterns but also on diverse and relevant topics.
We observe that a good question is a natural composition of {\it interrogatives}, {\it topic words}, and {\it ordinary words}. 
Interrogatives lexicalize the pattern of questioning, topic words address the key information for topic transition in dialogue, and ordinary words play syntactical and grammatical roles in making a natural sentence. 
We devise two typed decoders (\textit{soft typed decoder} and \textit{hard typed decoder}) in which a type distribution over the three types is estimated and used to modulate the final generation distribution.
Extensive experiments show that the typed decoders outperform state-of-the-art baselines and can generate more meaningful questions.
\end{abstract}

\section{Introduction}
Learning to ask questions (or, question generation) aims to generate a question to a given input. 
Deciding what to ask and how is an indicator of machine understanding~\cite{mostafazadeh2016generating}, as demonstrated in machine comprehension~\cite{du2017learning,zhou2017neural,yuan2017machine} and question answering~\cite{tang2017question,wang2017joint}.
Raising good questions is essential to conversational systems because a good system can well interact with users by asking and responding~\cite{li2016learning}. Furthermore, asking questions is one of the important proactive behaviors that can drive dialogues to go deeper and further~\cite{Yu2016strategy}.

Question generation (QG) in open-domain conversational systems differs substantially from the traditional QG tasks. 
The ultimate goal of this task is to enhance the {\it interactiveness and persistence of human-machine interactions}, while for traditional QG tasks, seeking information through a generated question is the major purpose. 
The response to a generated question will be supplied in the following conversations, which may be novel but not necessarily occur in the input as that in traditional QG ~\cite{du2017learning,yuan2017machine, tang2017question,wang2017joint, mostafazadeh2016generating}. Thus, the purpose of this task is to spark novel yet related information to drive the interactions to continue. 

Due to the different purposes, this task is unique in two aspects: 
it requires to question not only in various patterns but also about diverse yet relevant topics.
\textbf{First}, there are various questioning patterns for the same input, such as Yes-no questions and Wh-questions with different interrogatives. Diversified questioning patterns make dialogue interactions richer and more flexible. Instead, traditional QG tasks can be roughly addressed by syntactic transformation~\cite{Andrenucci2005Automated,popowich2013generating}, or implicitly modeled by neural models~ \cite{du2017learning}. In such tasks, the information questioned on is pre-specified and usually determines the pattern of questioning. For instance, asking Who-question for a given person, or Where-question for a given location.    

\textbf{Second}, this task requires to address much more transitional topics of a given input, which is the nature of conversational systems. For instance, for the input {\it ``I went to dinner with my friends"}, we may question about topics such as {\it friend, cuisine, price, place} and \textit{taste}. Thus, this task generally requires \textit{scene understanding} to imagine and comprehend a scenario (e.g., {\it dining at a restaurant}) that can be interpreted by topics related to the input. However, in traditional QG tasks, the core information to be questioned on is pre-specified and rather static, and {\it paraphrasing} is more required. 

\begin{figure}[!ht]
    \centering
    \includegraphics[width=3.0in]{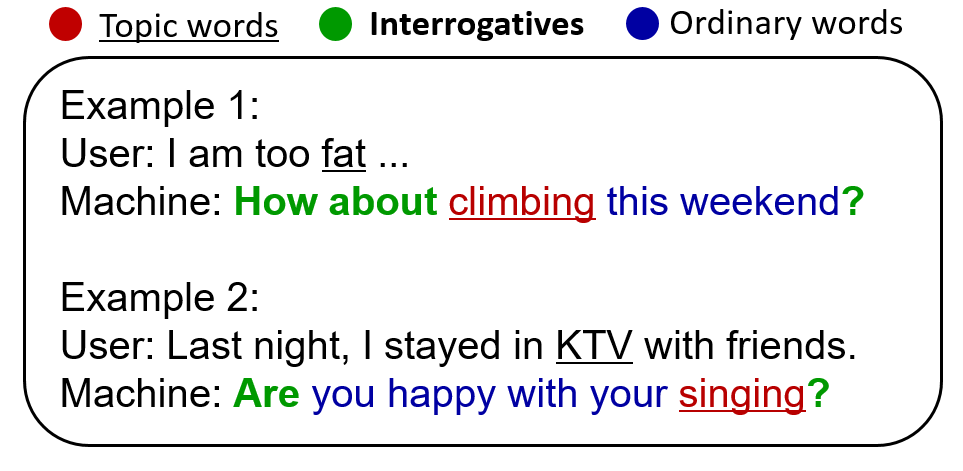}
    \caption{Good questions in conversational systems are a natural composition of interrogatives, topic words, and ordinary words.}
    \label{example1}
    \end{figure}
    
Undoubtedly, asking good questions in conversational systems needs to address the above issues (\textit{questioning with diversified patterns, and addressing transitional topics naturally in a generated question}). As shown in Figure \ref{example1}, a good question is a natural composition of interrogatives, topic words, and ordinary words. Interrogatives indicate the pattern of questioning, topic words address the key information of topic transition, and ordinary words play syntactical and grammatical roles in making a natural sentence.   

We thus classify the words in a question into three types: \textbf{interrogative}, \textbf{topic word}, and \textbf{ordinary word} automatically. We then devise two decoders, Soft Typed Decoder (STD) and Hard Typed Decoder (HTD), for question generation in conversational systems\footnote{To simplify the task, as a preliminary research, we consider the one-round conversational system.}. STD deals with word types in a latent and implicit manner, while HTD in a more explicit way. At each decoding position, we firstly estimate a type distribution over word types. STD applies a mixture of type-specific generation distributions where type probabilities are the coefficients. By contrast, HTD reshapes the type distribution by Gumbel-softmax and modulates the generation distribution by type probabilities. 
Our contributions are as follows:
\begin{itemize}

\item{To the best of our knowledge, this is the first study on question generation in the setting of conversational systems. We analyze the key differences between this new task and other traditional question generation tasks. }

\item{We devise soft and hard typed decoders to ask good questions by capturing different roles of different word types. Such typed decoders may be applicable to other generation tasks if word semantic types can be identified.}

\end{itemize}

\section{Related Work}
Traditional question generation can be seen in task-oriented dialogue system~\cite{Curto2012Question}, sentence transformation \cite{vanderwende2008importance}, machine comprehension~\cite{du2017learning,zhou2017neural,yuan2017machine,Subramanian2017Neural}, question answering~\cite{qin2015question,tang2017question,wang2017joint,Song2017A}, and visual question answering~\cite{mostafazadeh2016generating}. 
In such tasks, the answer is known and is part of the input to the generated question. Meanwhile, the generation tasks are not required to predict additional topics since all the information has been provided in the input. They are applicable in scenarios such as designing questions for reading comprehension~\cite{du2017learning,zhou2017mechanism,yuan2017machine}, and justifying the visual understanding by generating questions to a given image (video) ~\cite{mostafazadeh2016generating}. 

In general, traditional QG tasks can be addressed by the heuristic rule-based reordering methods \cite{Andrenucci2005Automated,ali2010automation,Heilman2010Good}, slot-filling with question templates \cite{popowich2013generating,chali2016ranking,labutov2015deep}, or implicitly modeled by recent neural models\cite{du2017learning,zhou2017neural,yuan2017machine,Song2017A,Subramanian2017Neural}. These tasks generally do not require to generate a question with various patterns: for a given answer and a supporting text, the question type is usually decided by the input.

Question generation in large-scale, open-domain dialogue systems is relatively unexplored. \citeauthor{li2016learning} \shortcite{li2016learning} showed that asking questions in task-oriented dialogues can offer useful feedback to facilitate learning through interactions. Several questioning mechanisms were devised with hand-crafted templates, but unfortunately not applicable to open-domain conversational systems. Similar to our goal, a visual QG task is proposed to generate a question to interact with other people, given an image as input \cite{mostafazadeh2016generating}.  

\begin{figure*}[t]
\centering
\includegraphics[width=6in]{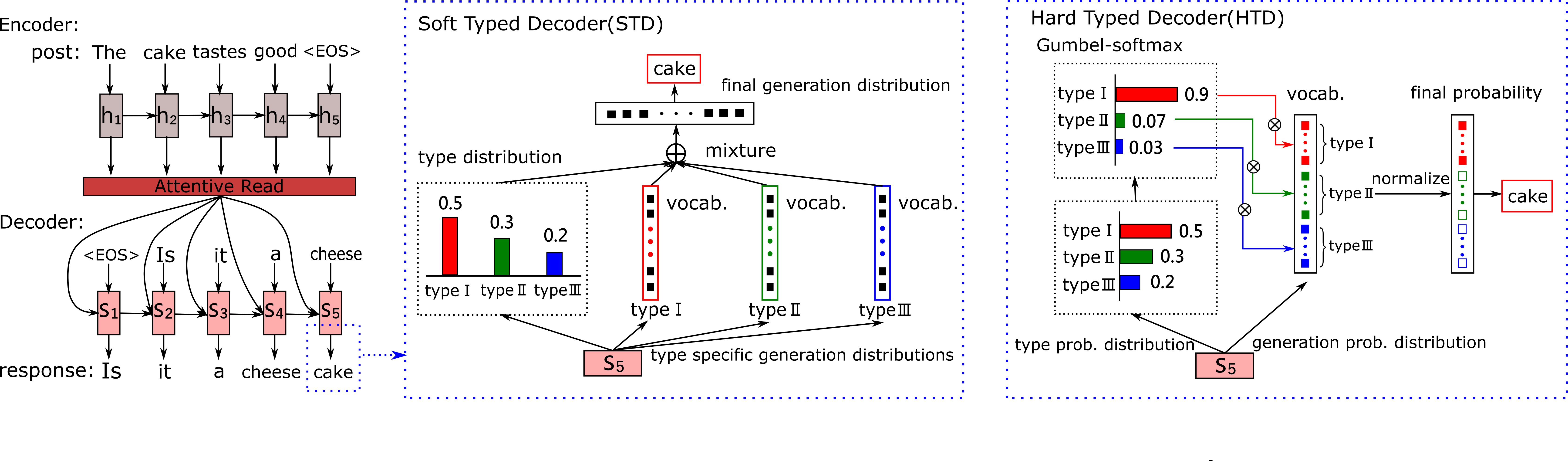}
\caption{Illustration of STD and HTD. STD applies a mixture of type-specific generation distributions where type probabilities are the coefficients. In HTD, the type probability distribution is reshaped by Gumbel-softmax and then used to modulate the generation distribution. In STD, the generation distribution is over the same vocabulary whereas dynamic     vocabularies are applied in HTD.}
\label{fig_model1}
\end{figure*}

\section{Methodology}
\subsection{Overview}
    The task of question generation in conversational systems can be formalized as follows: given a user post $X=x_1x_2\cdots x_m$, the system should generate a natural and meaningful question $Y=y_1y_2\cdots y_n$ to interact with the user, formally as
    \begin{align}
        {Y^*} = \mathop{argmax}\limits_{Y} \mathcal{P}(Y|X). \nonumber
    \end{align}

As aforementioned, asking good questions in conversational systems requires to question with diversified patterns and address transitional topics naturally in a question.  
To this end, we classify the words in a sentence into three types: {\it interrogative, topic word}, and {\it ordinary word}, as shown in Figure \ref{example1}.
During training, the type of each word in a question is decided automatically\footnote{Though there may be errors in word type classification, we found it works well in response generation.}. We manually collected about 20 interrogatives. The verbs and nouns in a question are treated as topic words, and all the other words as ordinary words.
During test, we resort to PMI~\cite{church1990word} to predict a few topic words for a given post.

On top of an encoder-decoder framework, we propose two decoders to effectively use word types in question generation. The first model is {\it soft typed decoder (STD)}. It estimates a  type distribution over word types and three type-specific generation distributions over the vocabulary, and then obtains a mixture of type-specific distributions for word generation. 

The second one is a {\it hard} form of STD, {\it hard typed decoder (HTD)},  in which we can control the decoding process more explicitly by approximating the operation of {\it argmax} with Gumbel-softmax \cite{jang2016categorical}. In both decoders, the final generation probability of a word is modulated by its word type. 

\subsection{Encoder-Decoder Framework}

Our model is based on the general encoder-decoder framework \cite{cho2014learning,sutskever2014sequence}. Formally, the model encodes an input sequence $X=x_1x_2\cdots x_m$ into a sequence of hidden states $\textbf{h}_i$, as follows,
        \begin{align}
            \textbf{h}_{t} &= \mathbf{GRU}(\textbf{h}_{t-1}, \bm{e}(x_t)),\nonumber
        \end{align}
where GRU denotes gated recurrent units~\cite{cho2014learning}, and $\bm{e}(x)$ is the word vector of word $x$.
The decoder generates a word sequence by sampling from the probability  $\mathcal{P}(y_t|y_{<t},X)$ ($y_{<t}=y_1y_2\cdots y_{t-1}$, the generated subsequence) which can be computed via
        \begin{align}
            &\mathcal{P}(y_t|y_{<t}, X) = \mathbf{MLP}(\textbf{s}_{ t}, \bm{e}(y_{t-1}), \textbf{c}_t), \nonumber \\  
            &\textbf{s}_{ t} = \mathbf{GRU}(\textbf{s}_{ t-1}, \bm{e}(y_{t-1}), \textbf{c}_t), \nonumber
        \end{align}
where $\textbf{s}_t$ is the state of the decoder at the time step $t$, and this GRU has different parameters with the one of the encoder. The context vector $\textbf{c}_t$ is an attentive read of the hidden states of the encoder as $\textbf{c}_t = \sum_{i=1}^T\alpha_{t,i}\textbf{h}_{i}$,
where the weight $\alpha_{t,i}$ is scored by another $\mathbf{MLP}(\textbf{s}_{ t-1}$, $\textbf{h}_{i})$ network.

\subsection{Soft Typed Decoder (STD)}
In a general encoder-decoder model, the decoder tends to generate universal, meaningless questions like ``{\it What's up?}" and ``{\it So what?}". In order to generate more meaningful questions, we propose a soft typed decoder. It assumes that each word has a \textit{latent type} among the set \{{\it interrogative, topic word, ordinary word}\}. The soft typed decoder firstly estimates a word type distribution over latent types in the given context, and then computes type-specific generation distributions over the entire vocabulary for different word types.
The final probability of generating a word is a mixture of type-specific generation distributions where the coefficients are type probabilities.

The final generation distribution $\mathcal{P}(y_t | y_{<t}, X)$ from which a word can be sampled, is given by
\begin{small}
\begin{align}
    &\mathcal{P}(y_t | y_{<t}, X) =   \nonumber\\ 
    &\sum_{i=1}^k \mathcal{P}(y_t | ty_t=c_i, y_{<t}, X) \cdot \mathcal{P}(ty_t=c_i | y_{<t}, X),
    \label{eq-mixture}
\end{align}
\end{small}
where $ty_t$ denotes the word type at time step $t$ and $c_i$ is a word type.
Apparently, this formulation states that the final generation probability is a mixture of the type-specific generation probabilities $\mathcal{P}(y_t | ty_t=c_i, y_{<t}, X)$, weighted by the probability of the type distribution $\mathcal{P}(ty_t = c_i | y_{<t}, {X})$. We name this decoder as \textit{soft typed decoder}. In this model, word type is latent because we do not need to specify the type of a word explicitly. In other words, each word can belong to any of the three types, but with different probabilities given the current context.

 The probability distribution over word types $\mathcal{C}=\{c_1,c_2,\cdots,c_k\}$ ($k=3$ in this paper) (termed as \textit{type distribution}) is given by
\begin{small}
\begin{align}
    \mathcal{P}(ty_t | y_{<t}, X) = softmax(\textbf{W}_{0} \textbf{s}_t + \textbf{b}_{0}), 
    \label{eq-type-prob}
\end{align}
\end{small}
where $s_t$ is the hidden state of the decoder at time step $t$, $\textbf{W}_{0} \in R^{k\times d}$, and $d$ is the dimension of the hidden state.

The type-specific generation distribution is given by
\begin{small}
\begin{align}
    \mathcal{P}(y_t | ty_t=c_i, y_{<t}, X) = softmax(\textbf{W}_{c_i} \textbf{s}_t + \textbf{b}_{c_i}), \nonumber
\end{align}
\end{small}
where $\textbf{W}_{c_i} \in R^{|V|\times d}$ and $|V|$ is the size of the entire vocabulary.
Note that the type-specific generation distribution is parameterized by $\textbf{W}_{c_i}$, indicating that the distribution for each word type has its own parameters. 

Instead of using a single distribution $\mathcal{P}(y_t|y_{<t},X)$ as in a general Seq2Seq decoder, our soft typed decoder enriches the model by applying multiple type-specific generation distributions. This enables the model to express more information about the next word to be generated. Also note that the generation distribution is over the same vocabulary, and therefore there is no need to specify word types explicitly.


\subsection{Hard Typed Decoder (HTD)}

In the soft typed decoder, we assume that each word is a distribution over the word types. In this sense, the type of a word is \textbf{implicit}. We do not need to specify the type of each word \textbf{explicitly}. 
In the hard typed decoder, words in the entire vocabulary are dynamically classified into three types for each post, and the decoder first estimates a type distribution at each position and then generates a word with the highest type probability.  
This process can be formulated as follows:
\begin{align}
    &c^* = \mathop{\arg\max}_{c_i} \mathcal{P}(ty_t=c_i | y_{<t}, X), \label{c_star}
\\
    &\mathcal{P}(y_t | y_{<t}, X) = \mathcal{P}(y_t | ty_t=c^*, y_{<t},X).
\end{align}
This is essentially the \textit{hard} form of Eq. \ref{eq-mixture}, which just selects the type with the maximal probability. 
However, this {\it argmax} process may cause two problems. First, such a cascaded decision process (firstly selecting the most probable word type and secondly choosing a word from that type) may lead to severe grammatical errors if the first selection is wrong. Second, {\it argmax} is discrete and non-differentiable, and it breaks the back-propagation path during training.

To make best use of word types in {\it hard typed decoder}, we address the above issues by applying {\it Gumbel-Softmax} \cite{jang2016categorical} to approximate the operation of {\it argmax}.
There are several steps in the decoder (see Figure \ref{fig_model1}):

\textbf{First},  the type of each word ({\it interrogative, topic, or ordinary}) in a question is decided automatically during training, as aforementioned.

\textbf{Second}, the generation probability distribution is estimated as usual,
\begin{align}
    \mathcal{P}(y_t | y_{<t}, X) = softmax(\textbf{W}_{0} \textbf{s}_t + \textbf{b}_{0}).
    \label{htd-prob}
\end{align}

Further, the type probability distribution at each decoding position is estimated as follows,
\begin{align}
    \mathcal{P}(ty_t | y_{<t}, X) = softmax(\textbf{W}_{1} \textbf{s}_t + \textbf{b}_{1}).
    \label{htd-type-prob}
\end{align}

\textbf{Third}, the generation probability for each word is modulated by its corresponding type probability:
        \begin{align}
        \mathcal{P}'(y_t | y_{<t}, X) = \mathcal{P}(y_t | y_{<t}, X)\cdot &\bm{m}({y_t}),  \nonumber\\
            \bm{m}(y_t)=
            \begin{cases}
                1&,c(y_t)= c^{*}\\
                0&,c(y_t)\not= c^{*}
            \end{cases}
            \label{gate_HTD}
        \end{align}
where $c(y_t)$ looks up the word type of word $y_t$, and $c^{*}$ is the type with the highest probability as defined in Eq. \ref{c_star}. This formulation has exactly the effect of {\it argmax}, where the decoder will only generate words of type with the highest probability.

To make $\mathcal{P}^*(y_t | y_{<t}, X)$ a distribution, we normalize these values by a normalization factor $Z$:
\begin{equation}
    Z = \frac{1}{\sum_{y_t\in \mathcal{V}}\mathcal{P}'(y_t|y_{<t},X)} \nonumber
\end{equation}
where $\mathcal{V}$ is the decoding vocabulary. Then, the final probability can be denoted by
\begin{equation}
    \mathcal{P}^*(y_t | y_{<t}, X) = Z\cdot \mathcal{P}'(y_t | y_{<t}, X).
    \label{p_star}
\end{equation}

As mentioned, in order to have an effect of {\it argmax} but still maintain the differentiability, we resort to {\it Gumbel-Softmax} \cite{jang2016categorical}, which is a differentiable surrogate to the {\it argmax} function. The type probability distribution is then adjusted to the following form:
\begin{align}
    &\bm{m}(y_t) = \mathbf{GS}(\mathcal{P}(ty_t = c(y_t) | y_{<t},X)),\nonumber \\
    &\mathbf{GS}(\pi_i) = \frac{e^{(log(\pi_i) + g_i)/\tau}}{\sum_{j=1}^k e^{(log(\pi_j) + g_j)/\tau}}, \label{gsoftmax}
\end{align}
where $\pi_1, \pi_2, \cdots, \pi_k$ represents the probabilities of the original categorical distribution, $g_j$ are i.i.d samples drawn from Gumbel(0,1)\footnote{If $u\sim Uniform(0,1)$, then $g=-log(-log(u))\sim Gumbel(0,1)$.}
    and $\tau$ is a constant that controls the smoothness of the distribution. When $\tau\rightarrow0$, Gumbel-Softmax performs like argmax, while if $\tau\rightarrow\infty$, Gumbel-Softmax performs like a uniform distribution. In our experiments, we set $\tau$ a constant between 0 and 1, making Gumbel-Softmax smoother than argmax, but sharper than normal softmax.

Note that in HTD, we apply dynamic vocabularies for different responses during training. The words in a response are classified into the three types dynamically. A specific type probability will only affect the words of that type. During test, for each post, topic words are predicted with PMI, interrogatives are picked from a small dictionary, and the rest of words in the vocabulary are treated as ordinary words.

\subsection{Loss Function}
    We adopt negative data likelihood (equivalent to cross entropy) as the loss function, and additionally, we apply supervision on the mixture weights of word types, formally as follows:
\begin{align}
    \Phi_1 &= \sum_t - \log \mathcal{P}(y_t=\tilde{y}_t|y_{<t}, X),\label{loss-part1} \\
    \Phi_2 &= \sum_t - \log \mathcal{P}(ty_t=\widetilde{ty}_t|y_{<t}, X),\\ 
    \Phi &= \Phi_1 + \lambda \Phi_2 
    \label{loss1},
\end{align}
   where $\widetilde{ty}_t$ represents the reference word type and $\tilde{y}_t$ represents the reference word at time $t$. $\lambda$ is a factor to balance the two loss terms, and we set $\lambda$=0.8 in our experiments.

    Note that for HTD, we substitute $\mathcal{P}^*(y_t=w_j |  y_{<t}, X)$ (as defined by Eq. \ref{p_star}) into Eq. \ref{loss-part1}.

\subsection{Topic Word Prediction}
    The only difference between training and inference is the means of choosing topic words. During training, we identify the nouns and verbs in a response as topic words; whereas during inference, we adopt PMI \cite{church1990word} and $Rel(k_i, X)$ to predict a set of topic words $k_i$ for an input post $X$, as defined below:
    \begin{align}
        &PMI(w_x, w_y) = log\frac{p(w_x, w_y)}{p_1(w_x) * p_2(w_y)},  \nonumber
        \\ 
        &Rel(k_i, X) = \sum_{w_x\in X} e^{PMI(w_x, k_i)}, \nonumber
        \label{relevance-score}
    \end{align}
    where $p_1(w)$/$p_2(w)$ represent the probability of word $w$ occurring in a post/response, respectively, and $p(w_x, w_y)$ is the probability of word $w_x$ occurring in a post and $w_y$ in a response.

During inference, we predict at most $20$ topic words for an input post. Too few words will affect the grammaticality since the predicted set contains infrequent topic words, while too many words introduce more common topics leading to more general responses.

\section{Experiment}
    \subsection{Dataset}
    To estimate the probabilities in PMI, we collected about 9 million post-response pairs from Weibo.
    To train our question generation models, we distilled the pairs whereby the responses are in question form with the help of around 20 hand-crafted templates. The templates contain a list of interrogatives and other implicit questioning patterns. Such patterns detect sentences led by words like \textit{what, how many, how about} or sentences ended with a \textit{question mark}.
    After that, we removed the pairs whose responses are universal questions that can be used to reply many different posts. This is a simple yet effective way to avoid situations where the type probability distribution is dominated by interrogatives and ordinary words.

Ultimately, we obtained the dataset comprising about 491,000 post-response pairs. We randomly selected 5,000 pairs for testing and another 5,000 for validation. The average number of words in post/response is 8.3/9.3 respectively. The dataset contains 66,547 different words, and 18,717 words appear more than 10 times. The dataset is available at: \url{http://coai.cs.tsinghua.edu.cn/hml/dataset/}. 

    \subsection{Baselines}
    We compared the proposed decoders with four state-of-the-art baselines.
    \\
    \textbf{Seq2Seq}: A simple encoder-decoder with attention mechanisms~\cite{luong2015effective}.
    \\ 
    \textbf{MA}: The mechanism-aware (MA) model applies multiple responding mechanisms represented by real-valued vectors ~\cite{zhou2017mechanism}. The number of mechanisms is set to $4$ and we randomly picked one response from the generated responses for evaluation to avoid selection bias. 
    \\ 
    \textbf{TA}: The {topic-aware (TA)} model generates informative responses by incorporating topic words predicted from the input post \cite{xing2017topic}.
    \\
    \textbf{ERM}: Elastic responding machine (ERM) adaptively selects a subset of responding mechanisms using reinforcement learning \cite{Zhou2018Elastic}. The settings are the same as the original paper.
\subsection{Experiment Settings}   
    Parameters were set as follows: we set the vocabulary size to $20,000$ and the dimension of word vectors as $100$. The word vectors were pre-trained with around 9 million post-response pairs from Weibo and were being updated during the training of the decoders. We applied the 4-layer GRU units (hidden states have 512 dimensions). These settings were also applied to all the baselines.
     $\lambda$ in Eq. \ref{loss1} is $0.8$.
    We set different values of $\tau$ in Gumbel-softmax at different stages of training. At the early stage, we set $\tau$ to a small value (0.6) to obtain a sharper reformed distribution (more like argmax). After several steps, we set $\tau$ to a larger value (0.8) to apply a more smoothing distribution. Our codes are available at: \url{https://github.com/victorywys/Learning2Ask_TypedDecoder}.

\subsection{Automatic Evaluation}
    We conducted automatic evaluation over the $5,000$ test posts. For each post, we obtained responses from the six models, and there are $30,000$ post-response pairs in total. 

\subsubsection{Evaluation Metrics} We adopted {\it perplexity} to quantify how well a model fits the data. Smaller values indicate better performance.
To evaluate the diversity of the responses, we employed {\it distinct-1} and {\it distinct-2} \cite{li2015diversity}. These two metrics calculates the proportion of the total number of distinct unigrams or bigrams to the total number of generated tokens in all the generated responses.

    Further, we calculated the proportion of the responses containing at least one topic word in the list predicted by PMI. This is to evaluate the ability of addressing topic words in response. We term this metric as {\it topical response ratio (TRR)}. We predicted 20 topic words with PMI for each post.

    \subsubsection{Results}
    Comparative results are presented in Table \ref{automatic}. STD and HTD perform fairly well with lower perplexities, higher distinct-1 and distinct-2 scores, and remarkably better topical response ratio (TRR). 
    Note that MA has the lowest perplexity because the model tends to generate more universal responses.
    \begin{table}[!ht]
    \small
    \centering
    \begin{tabular}{l c c c c}
    \toprule
    Model & Perplexity & Distinct-1 & Distinct-2 & TRR\\
    \midrule
    Seq2Seq & 63.71 & 0.0573 & 0.0836 & 6.6\% \\
    MA & \textbf{54.26} & 0.0576 & 0.0644 & 4.5\% \\
    TA & 58.89 & 0.1292 & 0.1781 & 8.7\% \\
    ERM & 67.62 & 0.0355 & 0.0710 & 4.5\% \\
    \hline 
    STD & 56.77 & 0.1325 & 0.2509 & 12.1\% \\
    HTD & 56.10 & \textbf{0.1875} & \textbf{0.3576} & \textbf{43.6\%} \\
    \bottomrule
    \end{tabular}
    \caption{Results of automatic evaluation.}
    \label{automatic}
    \end{table}

        \begin{table*}[!ht]
    \renewcommand{\arraystretch}{1.3}
        \small
        \centering
        \scalebox{0.9}{
        \begin{tabular}{l|ccc|ccc|ccc}
            \toprule
            \multirow{2}{*}{\centering Models} & \multicolumn{3}{c|}{Appropriateness} & \multicolumn{3}{c|}{Richness} & \multicolumn{3}{c}{Willingness} \\
            \cline{2-10}
            & Win (\%) & Lose (\%) & Tie (\%) & Win (\%) & Lose (\%) & Tie (\%) & Win (\%) & Lose (\%) & Tie (\%) \\
            \midrule
            STD vs. Seq2Seq & 42.0 & 38.6 & 19.4 & 37.2$^{**}$ & 15.2 & 47.6 & 45.4$^*$ & 38.6 & 16.0\\
            STD vs. MA & 39.6$^*$ & 31.2 & 29.2 & 32.6$^{**}$ & 16.8 & 50.6 & 49.4$^{**}$ & 27.0 & 23.6\\
            STD vs. TA & 42.2 & 40.0 & 17.8 & 49.0$^{**}$ & 5.4 & 45.6 & 47.6$^*$ & 40.2 & 12.2\\
            STD vs. ERM & 43.4$^*$ & 34.4 & 22.2 & 60.6$^{**}$ & 13.2 & 26.2 & 43.2$^*$ & 36.8 & 20.0\\
             
            \midrule
            HTD vs. Seq2Seq & 50.6$^{**}$ & 30.6 & 18.8 & 46.0$^{**}$ & 10.2 & 43.8 & 58.4$^{**}$ & 33.2 & 8.4\\
            HTD vs. MA & 54.8$^{**}$ & 24.4 & 20.8 & 45.0$^{**}$ & 17.0 & 38.0 & 67.0$^{**}$ & 18.0 & 15.0\\
            HTD vs. TA & 52.0$^{**}$ & 38.2 & 9.8 & 55.0$^{**}$ & 5.4 & 39.6 & 62.6$^{**}$ & 31.0 & 6.4\\
            HTD vs. ERM & 64.8$^{**}$ & 23.2 & 12.0 & 72.2$^{**}$ & 8.4 & 19.4 & 56.6$^{**}$ & 36.6 & 6.8 \\
            \midrule
            HTD vs. STD & 52.0$^{**}$ & 33.0 & 15.0 & 38.0$^{**}$ & 26.2 & 35.8 & 61.8$^{**}$ & 30.6 & 7.6\\
            \bottomrule
        \end{tabular}
        }
        \hspace{0.5cm}
        \caption{Annotation results. Win for ``A vs. B" means A is better than B. Significance tests with Z-test were conducted. Values marked with $^*$ means {\it p-value} $<$ 0.05, and $^{**}$ for {\it p-value} $<$ 0.01. }
        \label{manual_result}
    \end{table*}
    
    Our decoders have better distinct-1 and distinct-2 scores than baselines do, and HTD performs much better than the strongest baseline TA. Noticeably, the means of using topic information in our models differs substantially from that in TA. Our decoders predict whether a topic word should be decoded at each position, whereas TA takes as input topic word embeddings at all decoding positions. 
    
    Our decoders have remarkably better topic response ratios (TRR), indicating that they are more likely to include topic words in generation.

    \subsection{Manual Evaluation}
We resorted to a crowdsourcing service for manual annotation. 500 posts were sampled for manual annotation\footnote{During the sampling process, we removed those posts that are only interpretable with other context or background.}. We conducted pair-wise comparison between two responses generated by two models for the same post. In total, there are 4,500 pairs to be compared. For each response pair, five judges were hired to give a preference between the two responses, in terms of the following three metrics. Tie was allowed, and system identifiers were masked during annotation.
    
\subsubsection{Evaluation Metrics}
Each of the following metrics is evaluated independently on each pair-wise comparison:

\noindent    \textbf{Appropriateness}: measures whether a question is reasonable in logic and content, and whether it is questioning on the key information. Inappropriate questions are either irrelevant to the post, or have grammatical errors, or universal questions. 

\noindent    \textbf{Richness}: measures whether a response contains topic words that are relevant to a given post.
     
\noindent    \textbf{Willingness to respond}: measures whether a user will respond to a generated question. This metric is to justify how likely the generated questions can elicit further interactions. If people are willing to respond, the interactions can go further.
    
\subsubsection{Results}
 The label of each pair-wise comparison is decided by majority voting from five annotators.     Results shown in Table \ref{manual_result} indicate that STD and HTD outperform all the baselines in terms of all the metrics. 
 This demonstrates that our decoders produce more appropriate questions, with richer topics. Particularly, our decoders have substantially better {\it willingness scores}, indicating that questions generated by our models are more likely to elicit further interactions. 
 Noticeably, HTD outperforms STD significantly, indicating that it is beneficial to specify word types explicitly and apply dynamic vocabularies in generation. 
 
 We also observed that STD outperforms Seq2Seq and TA, but the differences are not significant in appropriateness. This is because STD generated about 7\% non-question responses which were judged as inappropriate, while Seq2Seq and TA generated universal questions (inappropriate too but beat STD in annotation) to these posts. 
\subsubsection{Annotation Statistics}    
The proportion of the pair-wise annotations in which at least three of five annotators assign the same label to a record is 90.57\%/93.11\%/96.62\% for appropriateness/ richness/willingness, respectively.
 The values show that we have fairly good agreements with majority voting.
 
\subsection{Questioning Pattern Distribution}
    To analyze whether the model can question with various patterns, we manually annotated the questioning patterns of the responses to 100 sampled posts. The patterns are classified into 11 types including Yes-No, How-, Why-, What-, When-, and Who- questions. We then calculated the KL divergence between the pattern type distribution by a model and that by human (i.e., gold responses).  

    Results in Table \ref{pattern_kl} show that the pattern distribution by our model is closer to that in human-written responses, indicating that our decoders can better learn questioning patterns from human language. Further investigation reveals that the baselines tend to generate simple questions like  {\it What?}(什么？) or {\it Really?}(真的吗), and constantly focus on using one or two question patterns whereas our decoders use more diversified patterns as appeared in the human language.

     \begin{table}[!ht]
    \small
    \centering
    \scalebox{0.85}{
    \begin{tabular}{c| c c c c | c c}  
   \toprule
    Model & Seq2Seq  & TA  & MA  & ERM & STD & HTD  \\
    \midrule
    KLD & 0.431 & 0.416 & 0.332 & 0.276 & \textbf{0.072} & 0.140\\
    \bottomrule
    \end{tabular}}
    \hspace{0.5cm}
    \caption{KL divergence between the questioning pattern distribution by a model and that by human.}
    \label{pattern_kl}
    \end{table}

\subsection{Examples of the Generated Questions}
     
Table \ref{generated-example} presents some generated questions by our decoders, which are more appropriate. On the contrary, Seq2Seq, MA and ERM tend to generate more universal questions. 
These examples also clearly show that asking questions in open-domain conversational systems requires {\it scene understanding}, which is verified by this scene example of {\it singing at karaoke}(在卡拉ok唱歌).

    \begin{table}[!ht]
    \begin{center}
    \small
    \scalebox{0.80}{
    \begin{tabular}{l m{6cm}}
    \toprule
    {\textbf{Post:}} & {我们昨天去\uline{唱歌}了} \\
     & {We \uline{{went to karaoke}} yesterday.}\\\midrule
     
    {\textbf{Seq2Seq:}} & \textbf{什么}时候\textbf{?} \\
     & \textbf{When?} \\
    {\textbf{MA:}}  & 你\textbf{怎么}知道我是在问\\
     & \textbf{How} did you know I am questioning you\textbf{?}\\
    {\textbf{TA:}}  & \textbf{什么?}\\
     & \textbf{What?} \\
    {\textbf{ERM:}} & \textbf{什么}情况\textbf{?}\\
     & \textbf{What} happened \textbf{?}\\
    {\textbf{STD:}}  & 去\textbf{哪里}\uline{{唱歌}}\textbf{?}\\
     & \textbf{Where} did you \uline{{sing karaoke}}\textbf{?} \\
    {\textbf{HTD:}}  &  你们\textbf{几}个人\uline{{唱歌}}\textbf{?}\\ 
     & \textbf{How many} people were \uline{{singing}} with you\textbf{?}\\\bottomrule
    \end{tabular}}
    \end{center}
    \caption{Generated question examples. Interrogative words are bolded and topic words underlined.}
    \label{generated-example}
    \end{table}
    
   \begin{table}[!ht]
    \begin{center}

    \scalebox{0.60}{
    \begin{tabular}{l p{9cm}}
    \toprule
    \textbf{Post-1:} & 不好不好。哼！\uline{{讨厌}}你啦。 \\ 
     & {No, no. Humph, I \uline{{hate}} you!} \\
    \textbf{Human:} & 我\textbf{怎么}了\textbf{?} \\ 
     & \textbf{What} is wrong with me\textbf{?} \\ 
    \textbf{STD:}  & 我\textbf{怎么}了\textbf{?} \\ 
     & \textbf{What} is wrong with me\textbf{?} \\
    \textbf{HTD:}  & 我\textbf{怎么}了\textbf{?} \\ 
     & \textbf{What} is wrong with me\textbf{?} \\ \midrule

    \textbf{Post-2:} & 刚用了\uline{杀虫剂}，现在已经有\uline{{效果}}了。 \\ 
     & {I just sprayed the \uline{insecticide}, and now the \uline{{effect}} appears.} \\ 
    \textbf{Human:} &  这么快!\uline{{效果}}\textbf{怎么}样\textbf{?}\\ 
     & So quick! \textbf{What} is the \uline{{effect}}\textbf{?} \\ 
    \textbf{STD:}  & \uline{{效果}}\textbf{如何?} \\ 
     & \textbf{What} is the \uline{{effect}}\textbf{?} \\ 
    \textbf{HTD:}  & \uline{{效果}}\textbf{如何?} \\ 
     & \textbf{What} is the \uline{{effect}}\textbf{?} \\ \midrule
    
    \textbf{Post-3:} & 我在\uline{{体育中心}}\uline{{上班}}。 \\ 
     & {I \uline{{work}} at the \uline{{sports center}}.} \\ 
    {\textbf{Human:}} & 你每天从\uline{{家}}\uline{{过去}}要\textbf{多久?} \\
     & \textbf{How} long does it \uline{{take}} you to \uline{{go}} there from \uline{{home}}\textbf{?} \\ 
    {\textbf{STD:}}  & 你是\textbf{什么}\uline{{部门}}的啊\textbf{?} \\ 
     & \textbf{Which} \uline{{department}} are you from\textbf{?} \\ 
    {\textbf{HTD:}}  & 那你是\uline{{体育学院}}的\uline{{学生}}\textbf{吗?} \\
     & \textbf{Are} you a \uline{{student}} from the \uline{{College of Physical Education}}\textbf{?} \\ \midrule

    {\textbf{Post-4:}} & 今天的\uline{{寿司}}真好吃。 \\ 
     & {The \uline{{sushi}} today is so delicious!} \\ 
    {\textbf{Human:}} & 我记得你不喜欢\uline{{吃}}\uline{{日式料理}}的不是\textbf{吗?} \\ 
     & I remember you didn't like \uline{{eating Japanese food}}, \textbf{did} you\textbf{?} \\ 
    {\textbf{STD:}}  & 你\textbf{什么}时候来\uline{{无锡}}，我\uline{{请}}你。 \\ 
     & \textbf{When} will you come to \uline{{Wuxi}}\textbf{?} It is my \uline{{treat}}. \\ 
    {\textbf{HTD:}}  & 是\textbf{吗?}改天\uline{{尝尝}}! \\ 
     & Really\textbf{?} I will \uline{{try}} sometime! \\ \bottomrule

    \end{tabular}}
    \end{center}
    \caption{Examples for typical questioning patterns. Interrogative words in response are bolded and topic words are underlined. }
    \label{question-pattern}
    \end{table}
To further understand how to ask good questions in conversational systems, we presented more generated question examples by human and our typed decoders.    
The first example (Post-1) in Table \ref{question-pattern} shows that a good question can be composed of only interrogatives and ordinary words. This shows the complexity of this task and also the necessity of modeling interrogatives. Post-2 indicates a typical questioning pattern which is questioning on a particular topic word (效果-{\it effect}) of the input. While for Post-3, the questions are asking about transitional topics of the input ({\it 上班-work $\rightarrow$ 部门-department; 体育中心-sports center $\rightarrow$ 体育学院-college of Physical Education}), indicating a typical case of {\bf topic transition} in our task (also seen in Post-4, 寿司-sushi $\rightarrow$日式料理-Japanese food). This example also demonstrates that for the same input, there are {\bf various questioning patterns}: a How-question asked by human, a Which-question by STD, and a Yes-No question by HTD. As for Post-4, the gold question requires a background that is only shared between the poster and responder, while STD and HTD tend to raise more general questions due to the lack of such shared knowledge.  
    
\subsection{Visualization of Type Distribution}    
\begin{figure}[!ht]
    \centering
    \includegraphics[width=3.0in]{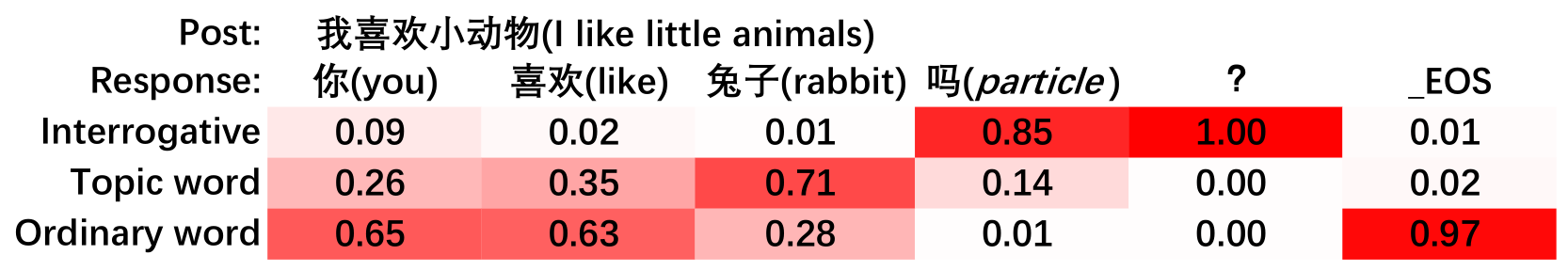}
    \caption{Type distribution examples from HTD. The generated question is ``你喜欢兔子吗？\textit{do you like rabbit?}". $\_EOS$ means end of sentence.}
    \label{fig_case2}
    \end{figure}

  To gain more insights into how a word type influence the generation process, we visualized the type probability at each decoding position in HTD. This example (Figure \ref{fig_case2}) shows that the model can capture word types well at different positions. For instance, at the first and second positions, ordinary words have the highest probabilities for generating 你-{\it you} and 喜欢-{\it like}, and at the third position, a topic word 兔子-{\it rabbit} is predicted while the last two positions are for interrogatives (a particle and a question mark).

\subsection{Error Analysis}
    We presented error type distribution by manually analyzing 100 bad responses sampled from STD and HTD respectively, where {\it bad} means the response by our model is worse than that by some baseline during the pair-wise annotation.

    There are 4 typical error types: {\it no topic words (NoT)} in a response (mainly universal questions), {\it wrong topics  (WrT)} where topic words are irrelevant, {\it type generation error (TGE)} where a wrong word type is predicted (See Eq. \ref{eq-type-prob}) and it causes grammatical errors, and {\it other errors}. 
    
    \begin{table}[!ht]
    \begin{center}
    \scalebox{0.77}{
    \begin{tabular}{m{2.5cm} c c c c}
    \toprule
    \textbf{Error Type} & \textbf{NoT} & \textbf{WrT}& \textbf{TGE} & \textbf{Others} \\ 
    \midrule
    STD & 34\% & 34\% & 29\% & 3\% \\
    HTD & 29\% & 39\% & 29\% & 3\% \\
    
    \bottomrule
    \end{tabular}}
    \end{center}
    \caption{Error type distribution.}
    \label{error_example}
    \end{table}
    The error distribution is shown in Table \ref{error_example}. For STD, most of the errors are attributed to no topic or wrong topics, while for HTD, the majority of errors fall into {\it wrong topics}. 
    
    \begin{CJK*}{UTF8}{gkai}
    \begin{table}[!ht]
    \begin{center}
    \small
    \scalebox{0.8}{
    \begin{tabular}{l l p{6cm}}
    \multirow{4}{*}{\textbf{~NoT}}\ldelim\{{4}{0.1pt} & {\textbf{Post-1:}} & {今天好开心啊!} \\
    & & {I am so happy today!}\\
    & {\textbf{STD:}}  & 你\textbf{怎么}知道\textbf{?}\\
    & & \textbf{How} do you know \textbf{?} \\ 
     
    \multirow{6}{*}{\textbf{WrT}}\ldelim\{{6}{0.1pt} & {\textbf{Post-2:}} & {\uline{{海报}}非常棒，期待若曦与我们男人的首度\uline{{合作}}。} \\
    & & {The \uline{{poster}} is great and we look forward to our first \uline{{cooperation}} with Ruoxi.}\\
    & {\textbf{HTD:}}  & 你\uline{{海报}}\textbf{怎么}样啊\textbf{?}\\
    & & \textbf{How} about your \uline{{poster}} \textbf{?} \\ 
     
     \multirow{7}{*}{\textbf{TGE}}\ldelim\{{7}{0.1pt} & {\textbf{Post-3:}} & {又\uline{{生病}}啦?\uline{{吃}}点\uline{{药}}就好了。} \\
    & & {\uline{{Got sick}} again? Just \uline{{take}} some \uline{{medicine}} and you'll be fine soon.}\\
    & {\textbf{STD:}} & 我也不知道\textbf{怎么}回事。\\
    & & I don't know \textbf{what} happened. \\ 
    & {\textbf{HTD:}}  & \uline{{肠胃痛}}了，\uline{{上火}}\textbf{吗?}\\
    & & \uline{{Stomach aching}}, \uline{{ulcer}} \textbf{?}
    \end{tabular}}
    \end{center}
    \caption{Cases for the error types with interrogative words bolded and topic words underlined.}
    \label{error-cases}
    \end{table}
\end{CJK*}
\begin{CJK*}{UTF8}{gkai}
    There are typical cases for these error types: 
    (1) Posts such as {\it ``I am so happy today!"} contains no topic words or rare topic words. In this case, our method is unable to predict the topic words so that the models tend to generate universal questions. 
    This happens more frequently in STD because the topic words are not specified explicitly.
    (2) Posts contains multiple topic words, but the model sometimes focuses on an inappropriate one. 
    For instance, for Post-2 in Table \ref{error-cases}, 
    HTD focused on {\it 海报-poster}
    but {\it 合作-cooperation} is a proper one to be focused on.
    (3) For complex posts, the models failed to predict the correct word type in response. For Post-3,
   STD generated a declarative sentence and HTD generated a question which, however, is not adequate within the context. 
\end{CJK*}    

These cases show that controlling the questioning patterns and the informativeness of the content faces with the compatibility issue, which is challenging in language generation. These errors are also partially due to the imperfect ability of topic word prediction by PMI, which is challenging itself in open-domain conversational systems.

\section{Conclusion and Future Work}
We present two typed decoders to generate questions in open-domain conversational systems. The decoders firstly estimate a type distribution over word types, and then use the type distribution to modulate the final word generation distribution. Through modeling the word types in language generation, the proposed decoders are able to question with various patterns and address novel yet related transitional topics in a generated question. Results show that our models can generate more appropriate questions, with richer topics, thereby more likely to elicit further interactions.

The work can be extended to multi-turn conversation generation by including an additional detector predicting when to ask a question. The detector can be implemented by a classifier or some heuristics. Furthermore, the typed decoders are applicable to the settings where word types can be easily obtained, such as in emotional text generation \cite{affect-lm,ecm}.

\section*{Acknowledgements}
    This work was partly supported by the National Science Foundation of China under grant No.61272227/61332007 and the National Basic Research Program (973 Program) under grant No. 2013CB329403. We would like to thank Prof. Xiaoyan Zhu for her generous support.

\bibliography{bib}
\bibliographystyle{acl_natbib}
\end{CJK*}

\end{document}